\documentclass[conference]{ieeeconf}
\IEEEoverridecommandlockouts
\usepackage{cite}
\usepackage{amsmath,amssymb,amsfonts}
\usepackage{algorithmic}
\usepackage{graphicx}
\usepackage{textcomp}
\usepackage{xcolor}
\usepackage{comment}
\usepackage{caption}
\usepackage{booktabs} 
\usepackage{siunitx}
\usepackage{censor}

\definecolor{squidred}{RGB}{229, 26, 26}
\definecolor{squidgreen}{RGB}{0, 163, 163}
\definecolor{squidpink}{RGB}{255, 0, 128}
\definecolor{squidyellow}{RGB}{228, 210, 57}

\def\BibTeX{{\rm B\kern-.05em{\sc i\kern-.025em b}\kern-.08em
    T\kern-.1667em\lower.7ex\hbox{E}\kern-.125emX}}

\begin{document}

\title{\Large \textbf{\textcolor{squidgreen}{Robot} \textcolor{squidpink}{Squid} \textcolor{squidyellow}{Game}}: Quadrupedal Locomotion for Traversing Narrow Tunnels
}

\author{Amir Hossain Raj, Dibyendu Das, and Xuesu Xiao\\
\thanks{All authors are with the Department of Computer Science, George Mason University
        {\tt\footnotesize \{araj20, ddas6, xiao\}@gmu.edu}}}

\maketitle

\begin{abstract}
Quadruped robots demonstrate exceptional potential for navigating complex terrain in critical applications such as search-and-rescue missions and infrastructure inspection. However, autonomous traversal of confined 3D environments—including tunnels, caves, and collapsed structures—remains a significant challenge. Existing methods often struggle with rigid gait patterns, limited adaptability to diverse geometries, and reliance on oversimplified environmental assumptions. This paper introduces a Reinforcement Learning (RL) framework that combines procedural environment generation with policy distillation to enable robust locomotion across various tunnel configurations. Our approach leverages a teacher-student training paradigm, where specialized expert policies trained on procedurally generated tunnel geometries transfer their knowledge to a unified student policy. This strategy eliminates the need for complex reward shaping in end-to-end RL training, simplifying the process by breaking down complicated tasks into smaller, more manageable components that are easier for the robot to learn. By synthesizing diverse tunnel structures during training and distilling navigation strategies into a generalizable policy, our method achieves consistent traversal across complex spatial constraints where conventional approaches fail. We demonstrate, through both simulation and real-world experiments, that our method enables quadruped robots to successfully traverse challenging, confined tunnel environments.
\end{abstract}

\section{Introduction}
The field of legged robotics has witnessed remarkable progress in recent years, with modern quadruped platforms demonstrating unprecedented agility across unstructured outdoor terrain \cite{SorokinOutdoor,caluwaerts2023barkourbenchmarkinganimallevelagility,Lee2020LearningQuadrupedal}. (e.g.,\cite{Kumar2021RMARM} demonstrating robust adaptation to sand, mud and other terrains). Much research has focused on enabling robots to navigate challenging ground conditions \cite{Duan2022, Duan2023, Agarwal2022EgocentricVision}, with approaches ranging from sensorized paws that identify terrain properties \cite{vangen2023terrainrecognitioncontactforce} to sophisticated control algorithms that maintain stability on uneven surfaces \cite{Zhu_2021}. End-to-end systems using egocentric vision have demonstrated impressive capabilities in traversing stairs, curbs and stepping stones \cite{Agarwal2022EgocentricVision}, while learning-based methods now enable locomotion across risky terrains with sparse footholds \cite{yu2024walkingterrainreconstructionlearning}. However, a critical capability gap persists in confined three-dimensional (3D) environments where spatial constraints impose 360° navigation challenges. Such scenarios demand not only ground-level obstacle negotiation but also precise coordination of body posture, limb articulation, and environmental awareness to avoid ceiling collisions and lateral obstructions. Applications ranging from mine shaft inspections to urban disaster response require robots to operate in tunnel-like spaces characterized by irregular cross-sections, tight turns, and limited visual accessibility—environments where current locomotion strategies frequently fail.
\begin{figure}
    \centering
    \includegraphics[trim=0 110 70 120, clip,width=1\columnwidth]{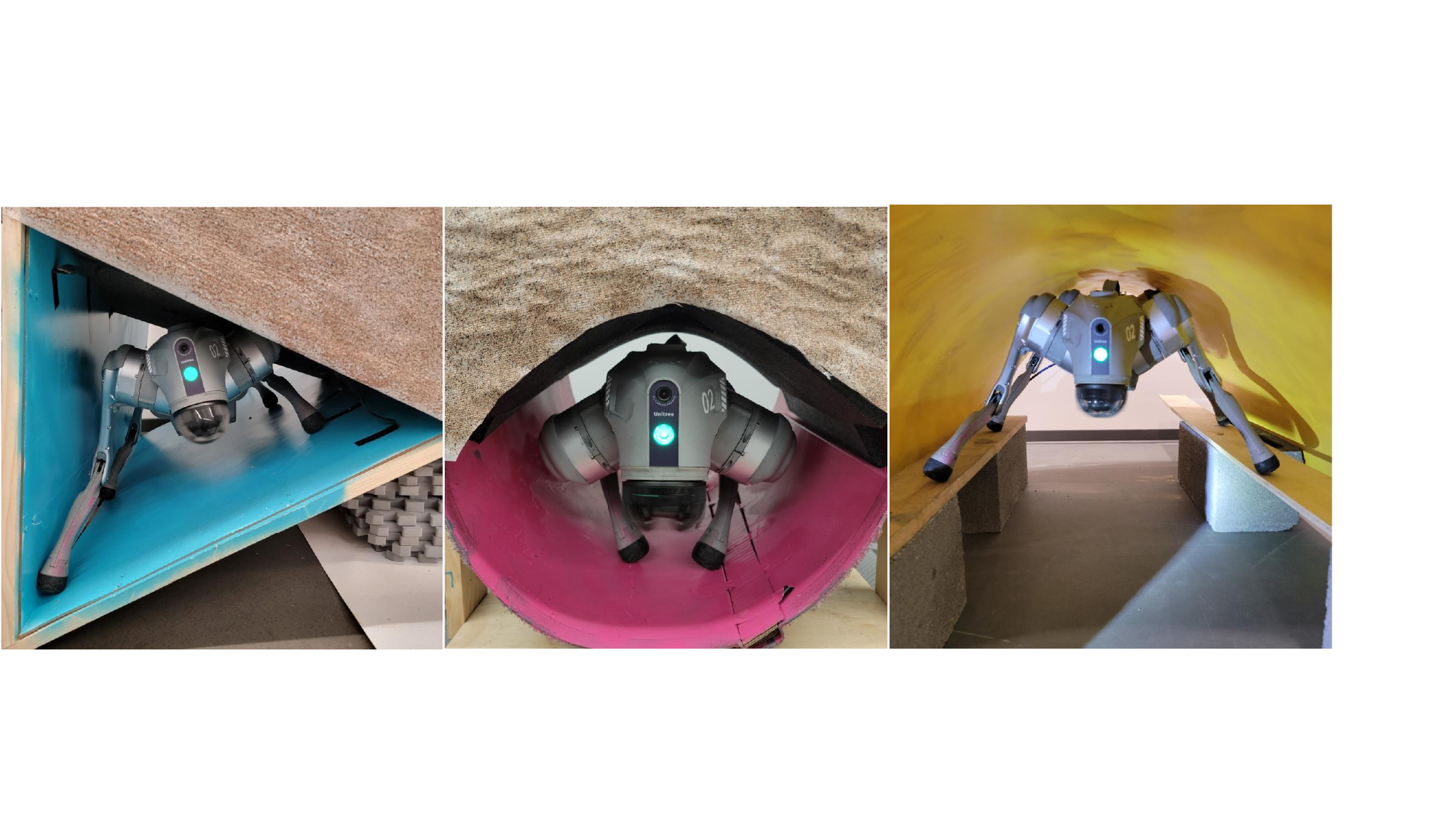}
    
\vspace{3pt}    \caption{\textsc{squid} is deployed in real-world tunnel environments, demonstrating the adaptability and robustness of the proposed approach. The quadrupedal robot relies on limited visual perception to navigate confined spaces, successfully traversing narrow passages and uneven terrain.}
    \label{fig::deployment}
    \vspace{-8pt}
\end{figure}

\begin{figure*}
    \centering
    \includegraphics[width=1\textwidth]{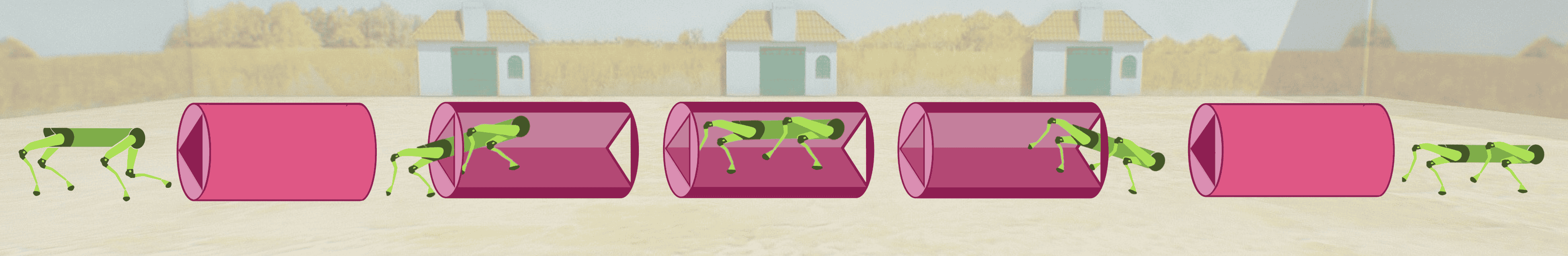}
    \vspace{-10pt}
    \caption{Quadruped robot executing its learned locomotion policy to traverse a confined tunnel, dynamically adjusting its posture to maintain stability and clearance.}
    \label{fig::animation}
    \vspace{-15pt}
\end{figure*}
Existing methods for navigating confined spaces primarily rely on either geometric planning with static gaits \cite{Buchanan2019WalkingPosture} or end-to-end Reinforcement Learning (RL) trained on simplified environmental models \cite{miki2024learning}. Although hierarchical frameworks that combine classical controllers and learned components appear promising, they encounter three core limitations \cite{Xu2024DexterousLocomotion}: (1) oversimplified training that fails to capture real-world structural diversity, (2) highly specialized policies that require substantial retuning for new tunnel shapes, and (3) sensitivity to sensory noise that undermines performance during deployment. Recent work by Buchanan et al. \cite{Buchanan2019WalkingPosture} demonstrates body-posture adaptation via a two-layer elevation map but remains constrained to predefined gait patterns, thereby failing to represent more diverse geometries (limitation~(1)) and necessitating specialized retuning (limitation~(2)). Meanwhile, RL-based approaches \cite{Xu2024DexterousLocomotion} can produce dynamic motions in cluttered environments yet remain tied to narrow training setups (limitation~(1)) and often degrade under sensor noise (limitation~(3)), preventing robust transfer to real-world scenarios.

In this paper, we present \textsc{squid} (Skill-fused Quadrupedal locomotion Using Imitation and Distillation), which leverages multi-expert learning and policy distillation  \cite{Rusu2015PolicyD} to enable robust traversal through confined tunnel environments.
\textsc{squid} addresses limitations (1), (2), and (3) described above by introducing a privileged learning framework that combines procedural environment generation with policy distillation.  Our approach is grounded in two central observations: first, that explicitly modeling the geometric variability of real-world confined 3D spaces (i.e., generating diverse tunnel geometries) is critical for overcoming oversimplified training setups (limitation~(1)). Second, decoupling perception-handling strategies from core locomotion is key to robust generalization—hence we initially train specialized “expert” policies with privileged information (focusing on locomotion), and later integrate perception as we transfer their expertise to a unified student policy. By doing so, we reduce specialization and the need for policy retuning (limitation~(2)) and mitigate sensitivity to noisy real-world sensors (limitation~(3)). Unlike prior work that trains a single policy on fixed obstacle distributions, our teacher–student architecture leverages multiple expert policies—each proficient in a distinct tunnel class—and distills their knowledge into a single model through supervised learning. This final policy can robustly handle a wide range of 3D constraints, effectively consolidating the locomotion skills of multiple experts while minimizing further environment-specific tuning.

The contributions of this work are threefold:

\begin{enumerate}
    \item Tunnel Simulation Pipeline: A procedural generation system creating 3D tunnel environments with parameterized geometric variations (cross-section asymmetry, slope transitions, etc.) that exceed the diversity of existing confined 3D training environments.
    \item Tunnel-Conditioned Policy Distillation: A multi-expert teacher--student distillation scheme that supervises the student with the tunnel-matched expert and uses class-balanced aggregation to retain locomotion behaviors across heterogeneous tunnel geometries within a single deployable policy.

    \item Real-World Depth-Based Deployment: Experimental validation showing that the distilled policy transfers to real hardware using only depth observations and proprioceptive inputs.
\end{enumerate}

\section{Related Work}
In the field of legged robotics, navigating confined spaces presents unique challenges that have been addressed through various methodologies. This section provides an overview of the existing literature, categorized into classical and hierarchical planning approaches, reinforcement learning techniques, privileged learning frameworks, and procedural environment generation methods.

\subsection{Classical and Hierarchical Planning Approaches} 

Early strategies for confined-space navigation relied on classical planning and optimization techniques. Buchanan et al. \cite{Buchanan2019WalkingPosture} introduced perceptive whole-body planning using elevation mapping and motion optimization to adapt robot posture in narrow environments. However, this approach depends on predefined motion primitives, which may not generalize well to irregular geometries. Similarly, Wellhausen et al. \cite{Wellhausen2023ArtPlanner} developed ArtPlanner, a sampling-based method employing reachability abstraction for legged robots operating in subterranean settings. This method necessitates handcrafted foothold safety heuristics, potentially limiting adaptability in dynamic terrains. Chestnutt et al. \cite{Chestnutt2009GlobalNavigation} proposed global navigation strategies using contact configuration graphs; however, the computational complexity of this approach poses challenges for real-time applications. 

\subsection{Reinforcement Learning for Confined-Space Locomotion}

Reinforcement Learning (RL) has emerged as a powerful tool for enhancing adaptability in unstructured environments. Xu et al. \cite{Xu2024DexterousLocomotion} proposed a hierarchical RL framework that combines classical waypoint planning with low-level policies for 360° obstacle avoidance, achieving successful real-world deployment in vertical shafts. However, reliance on explicit path planning can introduce coordination challenges between hierarchical layers. Miki et al. \cite{Miki2022RobustLocomotion} presented a two-level policy utilizing 3D volumetric representations to navigate under overhangs, though this approach requires separate terrain generators for each environment class. Rudin et al. \cite{Rudin2022Parkour} demonstrated end-to-end RL for dynamic skills such as leaping and crawling; however, their monolithic training framework faced difficulties in sustained confined navigation due to limited state memory. Additionally, approaches like Safe Locomotion within Confined Workspaces \cite{he2024agilesafelearningcollisionfree} using RL have been explored to enhance safety and adaptability in constrained environments. These studies underscore RL's potential but also highlight challenges in consolidating specialized skills across diverse geometries.

\subsection{Privileged Learning and Policy Distillation}

Privileged learning frameworks have been instrumental in improving simulation-to-reality transfer by decoupling perception and control. Hwangbo et al. \cite{Hwangbo2019LearningAgile} trained teacher policies with full state observability and subsequently distilled these navigation skills into vision-based student policies via behavioral cloning. Lee et al. \cite{Lee2020LearningQuadrupedal} extended this approach by incorporating curricular hindsight experience replay, facilitating fall recovery and high-speed terrain adaptation. Despite these advancements, single-policy architectures often struggle with conflicting skill requirements in multi-constraint spaces. Recent developments in policy distillation, such as Reinforcement Learning with Demonstrations and Guidance (RLDG), have shown that RL-generated training data can enhance the precision of generalist policies by 40\% over human demonstrations \cite{Chebotar2021ReinforcementLearning}. Chebotar et al. \cite{Chebotar2021ReinforcementLearning} demonstrated gradient-based distillation for unified control across manipulation tasks; however, this approach does not directly address the unique dynamics of legged locomotion in confined volumes.

\subsection{Procedural Environment Generation}

Procedural environment generation has become a cornerstone for training robust policies in diverse settings. Miki et al. \cite{miki2024learning} employed wave function collapse algorithms to synthesize confined spaces with overhangs, enhancing the training diversity for RL policies. Kumar et al. \cite{Kumar2021RMARM} developed automatic terrain difficulty curricula to promote agile locomotion, enabling robots to adapt to varying terrain complexities. Rudin et al. \cite{Rudin2022Parkour} created parkour courses featuring gaps and vertical obstacles; however, their parametric generators lacked the geometric diversity necessary for tunnel-like constraints. Our work advances this paradigm by introducing a physics-aware procedural tunnel generator capable of producing four distinct architectural classes (e.g., triangular, circular) with randomized dimensions, and slope transitions, thereby exceeding the variability present in prior datasets.

\section{Method}

Our framework combines procedural environment generation, privileged expert training, and vision-based policy distillation to enable robust quadrupedal navigation through confined tunnels. The system architecture progresses through three stages: (1) generating diverse tunnel geometries with parameterized difficulty levels, (2) training specialized expert policies for each tunnel class using privileged simulator information, and (3) distilling multiple experts into a single vision-based student policy through imitation learning.

\begin{table*}[h!]
\centering
\normalsize
% \caption{Tunnel Geometry Specifications}
\begin{tabular}{p{3cm}p{4cm}p{5cm}p{4cm}}
\toprule
\textbf{Tunnel Class} & \textbf{Cross-section Geometry} & \textbf{Difficulty Parameters} & \textbf{Generation Method} \\
\midrule
Triangle $\triangle$ & Three equal sides & Edge length $l_d = l_0 - 0.1d$, Rotation $\theta \sim U(0,360^\circ)$ & Random rotation per segment \\  
Circle $\bigcirc$ & Closed circular profile & Radius $r_d = r_0 - 0.05d$ & Fixed orientation \\  
Half-Circle $\bigcap$ & Semicircular arc & Radius $r_d = r_0 - 0.07d$, Z-normal $\in \{-1,+1\}$ (random) & Flipped Z-normal \\  
Gap $\vdash 	\dashv$ & Central gap with side shelves & Gap width $g_d = g_0 + 0.2d$, Shelf angle $\phi_d = \phi_0 + 5^\circ d$ & Symmetric shelves \\  
\bottomrule
\end{tabular}
\vspace{3pt}
\caption{Tunnel geometry specifications, detailing the cross-section geometry, difficulty parameters, and generation methods used for each tunnel class.}
\vspace{-15pt}
\label{tab:tunnel_params}
\end{table*}

\begin{figure}
    \centering
    \includegraphics[width=1\columnwidth]{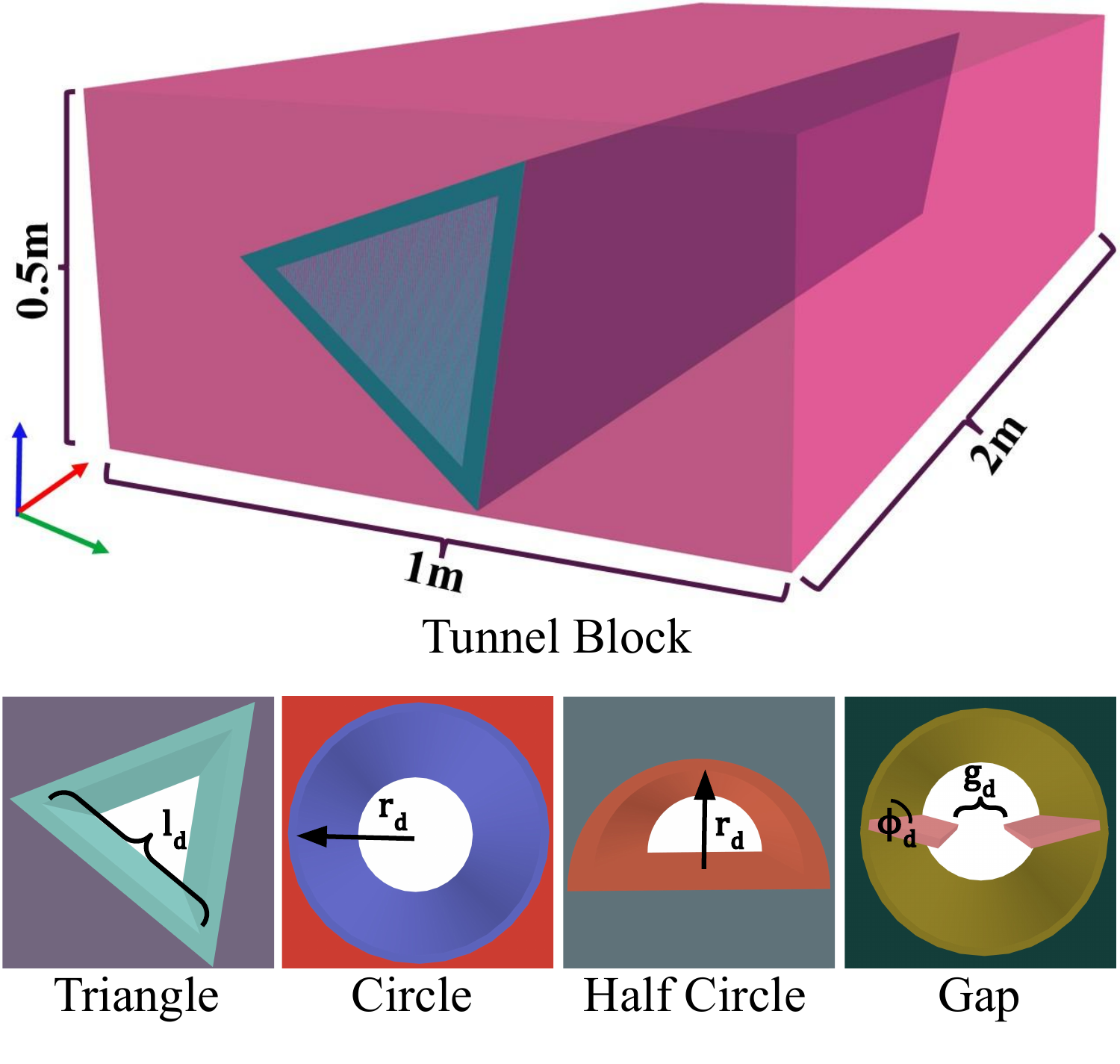}
    \caption{Simulation training environment for tunnel analysis. The upper portion of the image displays a tunnel block with its measurements and axis orientations. The lower section presents four distinct cross-sectional views of the tunnel, illustrating different structural variations used in the simulation.}
    \label{fig::tunnel_blocks}
    \vspace{-8pt}
\end{figure}

\subsection{Procedural Tunnel Generation}   

At the core of our approach is a procedural tunnel generation system designed to address the shortcomings of static and oversimplified training setups. Unlike traditional methods that train RL policies in fixed, predefined environments, our framework dynamically generates diverse tunnel configurations, exposing the robot to a broad range of spatial constraints. This ensures that the learned locomotion strategies remain adaptable rather than overfitting to a single geometry. The transitions in Fig. \ref{fig::animation} illustrate how the quadruped dynamically adjusts its posture as it moves through a tunnel, adapting to changes in spatial constraints.

  Each tunnel is constructed as a 3D block with a hollowed-out pathway for the robot to traverse (Fig. \ref{fig::tunnel_blocks}). To systematically vary the spatial constraints, we define four primary tunnel classes, each with tunable difficulty parameters (Table \ref{tab:tunnel_params}). These tunnel classes include: 
\begin{itemize}
    \item The equilateral triangle tunnel ($\triangle$) presents a sharp, angular interior that demands careful body rotation and frequent limb adjustments, with the available space shrinking as the edges shorten. 
    \item The full-circle tunnel ($\bigcirc$) offers a uniformly enclosed structure where reducing the radius progressively increases the difficulty, forcing the robot to crouch or adjust its gait dynamically.
    \item The half-circle tunnel ($\bigcap$) introduces additional complexity by randomly flipping its Z-normal, requiring adaptation to inverted terrain.
    \item The gap tunnel ($\vdash 	\dashv$) features a central void with elevated side shelves, posing a challenge for stable foothold selection, especially as the gap widens or the shelf angles increase.
\end{itemize}

Although these geometric primitives are simple, they capture several core geometric factors relevant to confined-space locomotion, including sharp angles, clearance variation, base inclination, and rotational asymmetry. They are intended to model fundamental spatial constraints rather than the full complexity of real-world confined environments. Additionally, the procedural nature of our framework prevents the learning process from being constrained to fixed obstacle distributions. Key environmental factors such as tunnel width, height, curvature, and inclination vary continuously across training episodes, ensuring a rich distribution of training data. By randomly rotating certain tunnels, flipping their orientations, and altering their connectivity, we force the robot to develop generalizable motion strategies rather than memorizing specific paths.

To further improve adaptability, our tunnel configurations are sequentially connected during training. The robot begins in simpler environments and gradually progresses to more complex ones, encountering increasingly tight, irregular, or asymmetric spaces. This curriculum-based approach ensures that the policy learns effective locomotion strategies incrementally, reinforcing fundamental movement principles before tackling highly constrained navigation.

By synthesizing a wide variety of tunnel structures and exposing the robot to continually changing spatial constraints, our method avoids the common pitfalls of static training environments. The result is a locomotion policy that retains the flexibility and robustness needed for real-world deployment, effectively bridging the gap between structured simulation training and unpredictable, confined 3D spaces.

\begin{figure}
    \centering
    \includegraphics[width=1\columnwidth]{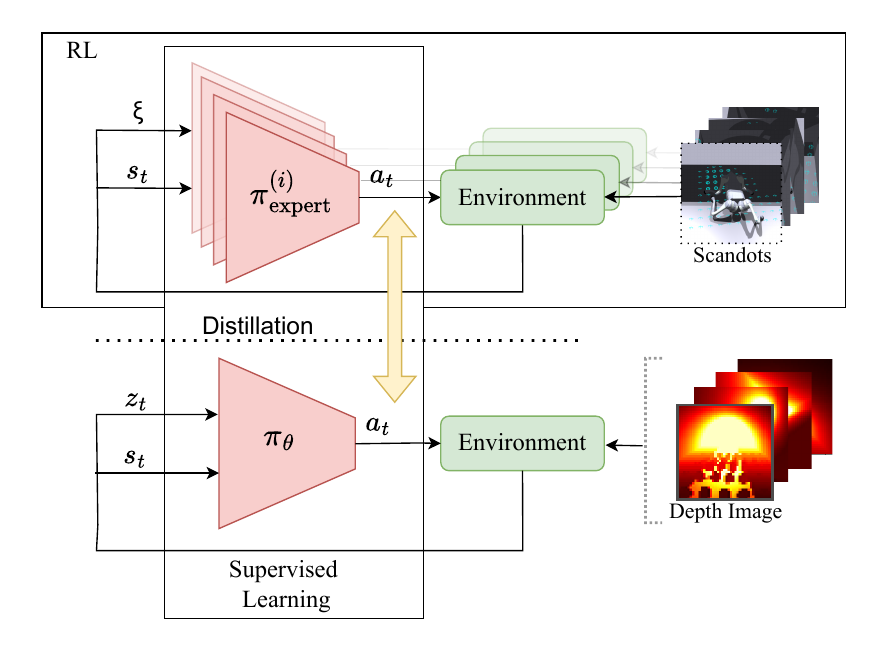} %\vspace{1px}
    \caption{Training pipeline for \textcolor{squidpink}{\textsc{squid}}. Teacher policies are first trained using RL with privileged information for each tunnel class. Distillation transfers expert knowledge to a unified student policy, which is trained using onboard sensing.}
    \label{fig::pipeline}
    \vspace{-8pt}
\end{figure}

\subsection{Privileged Teacher Policies}  

The teacher policies are trained using RL framework, mapping observations to actions that enable the robot to navigate tunnel environments of varying difficulty. Each teacher policy is trained independently within its assigned tunnel class, ensuring specialization in handling the unique constraints of that environment.
Given that we define four tunnel classes—triangle ($\triangle$), circle ($\bigcirc$), half-circle ($\bigcap$), and gap tunnel ($\vdash 	\dashv$)—we train four expert policies, one for each tunnel class. These experts leverage privileged state information during training to learn locomotion strategies, which are later distilled into a single deployable student policy.
To be specific, the $i$-th expert policy is denoted as:

\begin{equation}
    \pi_{\text{expert}}^{(i)}(a_t | s_t, \xi),
    \nonumber
\end{equation}
where, $s_t$ represents the state at time $t$, including proprioceptive and exteroceptive observations; $\xi$ denotes privileged simulator information, which includes ground-truth environmental details that are unavailable during student policy execution; and $a_t$ is the action taken at time $t$.

\subsubsection{Observation Space}

The observation space consists of proprioceptive and exteroceptive measurements that provide comprehensive information about the robot’s motion and surroundings. Specifically, in the observation space,
base linear and angular velocities ($\mathbf{v}_b, \boldsymbol{\omega}_b$) capture the robot’s movement dynamics; 
gravity vector orientation provides information about the robot’s pose relative to gravity; 
joint positions and velocities ($\mathbf{q}_j, \dot{\mathbf{q}}_j$) track the robot’s joint configurations;
previous actions maintain temporal consistency in decision-making; and
terrain measurement is a 108-dimensional grid around the robot’s base, encoding distances from the terrain surface to the robot’s body height.
By incorporating both proprioceptive and terrain-based sensory inputs, the teacher policies have access to high-fidelity state information, enabling them to learn robust locomotion strategies tailored to their specific tunnel class.

\subsubsection{Action Space}

Each expert policy outputs a 12-dimensional action vector, corresponding to desired joint positions for the 12 motors (three per leg). These actions are passed through a Proportional-Derivative (PD) controller, which converts them into motor torques for actuation:

\begin{equation}
    \boldsymbol{\tau}_j = k_p (\mathbf{q}_j^d - \mathbf{q}_j) + k_d (\dot{\mathbf{q}}_j^d - \dot{\mathbf{q}}_j),
    \nonumber
\end{equation}
where $\mathbf{q}_j^d$ and $\dot{\mathbf{q}}_j^d$ are the desired joint positions and velocities given by the teacher policy, $\mathbf{q}_j$ and $\dot{\mathbf{q}}_j$ are the current joint positions and velocities, and $k_p$ and $k_d$ are proportional and derivative gains controlling the system stiffness and damping.
Using joint position control instead of direct torque control ensures stable learning and efficient locomotion, as the system does not need to model complex actuator dynamics explicitly.

\subsubsection{Reward Function}

The reward function encourages efficient, stable, and collision-free locomotion while minimizing energy consumption. It is formulated as a weighted sum of individual reward terms (Table~\ref{tab:reward-terms}), where each term reinforces a specific desirable behavior.

To ensure precise trajectory tracking, we include linear velocity tracking ($r_{\text{lv}}$) and angular velocity tracking ($r_{\text{av}}$), which encourage the robot to match a target translational velocity $\mathbf{v}^*_{b,xy}$ and yaw rate $\omega^*_{b,z}$, respectively. These terms penalize deviations from the desired motion commands.
To maintain stable locomotion, vertical velocity ($r_{\text{vp}}$) and horizontal angular velocity penalties ($r_{\text{ap}}$) discourage excessive fluctuations in body movement. Additionally, joint motion ($r_{\text{jm}}$) and torque penalties ($r_{\tau}$) ensure smooth and efficient actuation by penalizing high joint accelerations, velocities, and large torque outputs.
Collision avoidance is enforced through a collision penalty ($r_{\text{coll}}$), which assigns negative rewards for contacts with tunnel walls, encouraging safer navigation through constrained spaces.
Finally, step duration reward ($r_{\text{step}}$) is introduced to promote structured footstep timing. It is based on the air time $t_{\text{air},f}$ of each leg $f$, ensuring a balance between stance and swing phases. A reference duration of 0.5 is used to encourage stable and efficient gaits.

\begin{table}[ht]
\begin{tabular}{p{1.5cm}p{3cm}p{3cm}}
  \toprule
  \textbf{Reward Term} & \textbf{Equation} & \textbf{Description} \\
  \midrule
  $r_{\text{lv}}$ & $\exp\left(-\frac{\|\mathbf{v}^*_{b,xy} - \mathbf{v}_{b,xy}\|^2}{0.25}\right)$ & Linear velocity tracking \\
  % \hline
  $r_{\text{av}}$ & $\exp\left(-\frac{\|\omega^*_{b,z} - \omega_{b,z}\|^2}{0.25}\right)$ & Angular velocity tracking \\
  % \hline
  $r_{\text{vp}}$ & $-v_{b,z}^2$ & Vertical velocity penalty \\
  % \hline
  $r_{\text{ap}}$ & $-\|\omega_{b,xy}\|^2$ & Horizontal angular velocity penalty \\
  % \hline
  $r_{\text{jm}}$ & $-\|\ddot{\mathbf{q}}_j\|^2 - \|\dot{\mathbf{q}}_j\|^2$ & Joint motion penalty \\
  % \hline
  $r_{\tau}$ & $-\|\boldsymbol{\tau}_j\|^2$ & Joint torque penalty \\
  % \hline
  $r_{\text{coll}}$ & $-n_{\text{collision}}$ & Collision penalty \\
  % \hline
  $r_{\text{step}}$ & $\displaystyle \sum_{f=1}^{4} \left(t_{\text{air},f} - 0.5\right)$ & Step duration reward \\
  \bottomrule
\end{tabular}
\vspace{3pt}
\caption{Reward Terms for Privileged Teacher Policies}
   \label{tab:reward-terms}
\end{table}

\subsection{Student Policy Using Distillation}

Once the four expert policies ($\pi_{\text{expert}}^{(i)}$) are trained, we employ a policy distillation framework to consolidate their knowledge into a single vision-based student policy (Fig.~\ref{fig::pipeline}). Unlike the experts, which rely on privileged simulator information, the student policy learns to navigate using depth images and historical proprioception, making it suitable for real-world deployment.

To achieve this, we use Dataset Aggregation (DAgger) \cite{ross2011reductionimitationlearningstructured}, an iterative imitation learning approach that mitigates distribution shift by collecting on-policy rollouts under the student policy while querying the expert policies for corrective supervision. The student policy is denoted as
\begin{equation}
    \pi_{\theta}(a_t \mid s_t, z_t),
    \nonumber
\end{equation}
where $s_t$ is the current proprioceptive state, $z_t$ represents the depth observation encoding environmental obstacles, and $a_t$ is the action controlling the robot’s motion.

Since each expert policy is trained for a specific tunnel class, we perform tunnel-conditioned distillation during student training. In particular, each training sample is associated with a tunnel class label $c_t \in \{1,2,3,4\}$ provided by the procedural environment generator, corresponding to triangle, circle, half-circle, and gap tunnels, respectively. The student is then supervised by the expert matched to that tunnel class.

The distillation objective is defined as
\begin{equation}
    L(\theta) = \mathbb{E}_{(s_t,z_t,c_t)\sim \mathcal{D}}
    \left[
    \ell\!\left(
    \pi_{\text{expert}}^{(c_t)}(s_t,\xi),\,
    \pi_{\theta}(s_t,z_t)
    \right)
    \right],
\end{equation}
Eq.~(1) defines the tunnel-conditioned distillation objective, where $\xi$ denotes the privileged simulator information available to the expert policies during training, $\ell(\cdot)$ is the action-matching loss, and $\mathcal{D}$ is the aggregated DAgger dataset collected under student rollouts.

To preserve behaviors across tunnel classes during training, we maintain a class-balanced replay buffer over the four tunnel types and sample balanced mini-batches during student updates. This reduces bias toward any single tunnel geometry and helps retain specialized behaviors from all expert policies within a single unified model.

\section{Experimental Results}

\begin{figure}
    \centering
    \includegraphics[width=0.7\columnwidth]{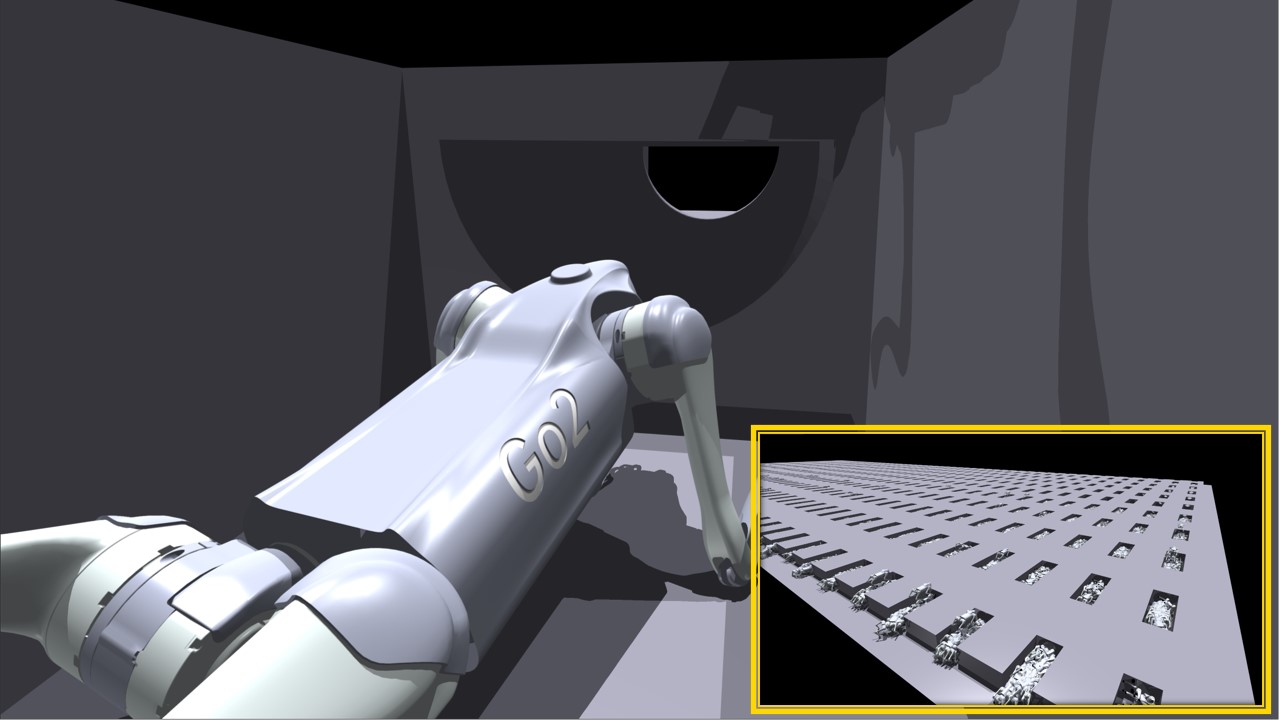}
    
\vspace{3pt}    \caption{Parallelized training of quadrupedal robots in confined tunnels (inset), with a zoomed-in view of a single environment.}
    \label{fig::training}
    \vspace{-8pt}
\end{figure}

In this section, we present an extensive evaluation of \textsc{squid} through simulation-based trials and real-world deployments. We first outline the experimental setup and performance metrics. We then compare our method against several baselines, conduct ablation studies to understand the contribution of each component, and finally discuss real-world test results.
\begin{figure*}
    \centering
    \includegraphics[width=1.8\columnwidth]{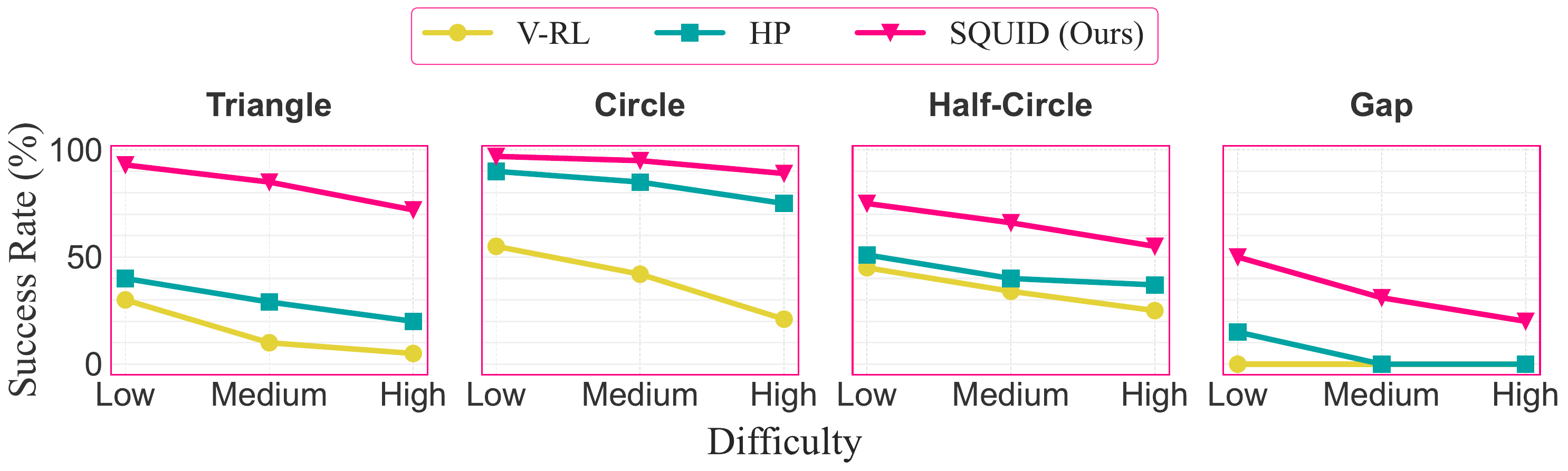}
    \vspace{2pt}
    \caption{Success Rate comparison across different tunnel classes and difficulty levels. Each plot represents a specific tunnel geometry, showing the performance of various methods (\textcolor{squidyellow}{V-RL}, \textcolor{squidgreen}{HP}, and \textcolor{squidpink}{\textsc{squid}}) as the difficulty increases. The \textsc{squid} policy consistently outperforms baselines, particularly in constrained environments like triangular and gap tunnels.}
    \label{fig::success_rate}
    % \vspace{-20pt}
\end{figure*}
\subsection{Experimental Setup and Metrics} 

\subsubsection{Robot Platform and Simulation} We use a simulated Unitree Go2 quadruped robot configured with 12 actuated joints. All simulation experiments are conducted using Isaac Gym \cite{makoviychuk2021isaacgymhighperformance}, allowing parallelized training and testing across multiple tunnel environments. The robot’s onboard sensor suite in the simulation includes forward-facing depth camera providing 64×48 pixel images; an Inertial Measurement Unit for orientation estimates; and joint encoders for proprioceptive feedback. We train and evaluate on four tunnel classes with increasing difficulty levels controlled by parameters in Table \ref{tab:tunnel_params} and values in Table \ref{tab:tunnel_difficulty}. Training is conducted using massive parallelization (Fig \ref{fig::training}) for each tunnel class where the columns are generated as a chain of tunnel blocks, with increasing difficulty, connected via narrow passages.

\subsubsection{Implementation and Training}
The teacher policies utilize a multi-layer perceptron architecture with tanh activation functions. These policies are trained using Proximal Policy Optimization (PPO) \cite{schulman2017proximalpolicyoptimizationalgorithms}. For the student policy, we employ a convolutional neural network~\cite{LeNet} encoder to process depth images, followed by a Gated Recurrent Unit \cite{Bahdanau2015Neural} to maintain temporal context across frames and proprioceptive history. The student network's final layers map encoded features to the 12-dimensional joint position action space. All training is conducted on an NVIDIA RTX 3090 GPU, enabling parallelization of 2048 environments in Isaac Gym. The teacher policies converge after approximately 10{,}000 iterations (\(\sim \)8 hours of training time per tunnel class). During distillation, depth images are augmented with random noise and occasional dropout to improve robustness against sensor imperfections in real-world deployment.

\subsubsection{Performance Metrics}  
We adopt four metrics reported as average over 50 trials per tunnel type and difficulty level: Success Rate is the percentage of trials in which the robot successfully reaches the tunnel exit without collisions that cause a reset; Trajectory Completion Time is the average time taken to traverse a tunnel segment; Collision Frequency is the number of body collisions with tunnel walls, measured per meter traveled; and Energy Consumption is computed as the accumulated actuator effort along the trajectory and normalized by the distance traveled, as defined in Eq.~(2).

\begin{equation}
E = \frac{1}{d} \sum_{t} \sum_{j=1}^{12} |\tau_{j,t}| \Delta t
\end{equation}

where $\tau_{j,t}$ is the torque applied at joint $j$ at time step $t$, $\Delta t$ is the control timestep, and $d$ is the distance traveled during the traversal.

\begin{table}[ht]
\centering
% \caption{Difficulty Parameters for Tunnel Classes}
% \label{tab:tunnel_difficulty}
\resizebox{\linewidth}{!}{%
\begin{tabular}{lccc}
\toprule
\textbf{Tunnel Class} & \textbf{Low} & \textbf{Medium} & \textbf{High} \\
\midrule
Triangle $\triangle$ & $l_d = [0.42m, 0.48m]$ & $l_d = [0.35m, 0.42m]$ & $l_d = [0.30m, 0.35m]$ \\  
& $\theta \sim U(0,90^\circ)$ & $\theta \sim U(0,180^\circ)$ & $\theta \sim U(0,360^\circ)$ \\  
\midrule
Circle $\bigcirc$ & $r_d = [0.22m, 0.24m]$ & $r_d = [0.16m, 0.22m]$ & $r_d =[0.12m, 0.18m]$ \\  
\midrule
Half-Circle $\bigcap$ & $r_d = [0.40m, 0.42m]$ & $r_d = [0.35m, 0.40m]$ & $r_d = [0.25m, 0.32m]$ \\  
  
\midrule
Gap $\vdash 	\dashv$ & $g_d = [0.1m, 0.2m]$ & $g_d = [0.2m, 0.35m]$ & $g_d = [0.35m, 0.4m]$ \\  
& $\phi_d = [0^\circ, 5^\circ$] & $\phi_d = [5^\circ, 10^\circ]$ & $\phi_d = [10^\circ, 15^\circ]$ \\  
\bottomrule
\end{tabular}
} % End resizebox
\vspace{3pt}
\caption{Difficulty Parameters for Tunnel Classes.}
\vspace{-8pt}
\label{tab:tunnel_difficulty}
\end{table}

\subsection{Comparison with Baselines}

We compare \textsc{squid} against two baselines in simulation using difficulty parameters from Table \ref{tab:tunnel_difficulty}. Vanilla RL (V-RL) is a single PPO agent trained end-to-end on all tunnel classes simultaneously with similar observation space as teacher policies. 
Hierarchical Planner (HP) \cite{Xu2024DexterousLocomotion} is a two-layer system where a high-level planner generates waypoints and a low-level controller executes footstep motions. The planner uses elevation mapping for obstacle detection.
 
\subsubsection{Success Rate}
Fig.~\ref{fig::success_rate} presents the Success Rates across tunnel classes and difficulty levels. \textsc{squid} achieves the highest Success Rates across all configurations with the most noticeable advantage in triangular and gap tunnels, where constrained navigation demands precise body articulation. HP performs well in structured tunnels (like Circle) but struggles in environments requiring adaptive motion strategies (like Gap). V-RL suffers from poor generalization, failing frequently in complex geometries and high difficulty levels.

\begin{figure}
    \centering
    \includegraphics[width=1\columnwidth]{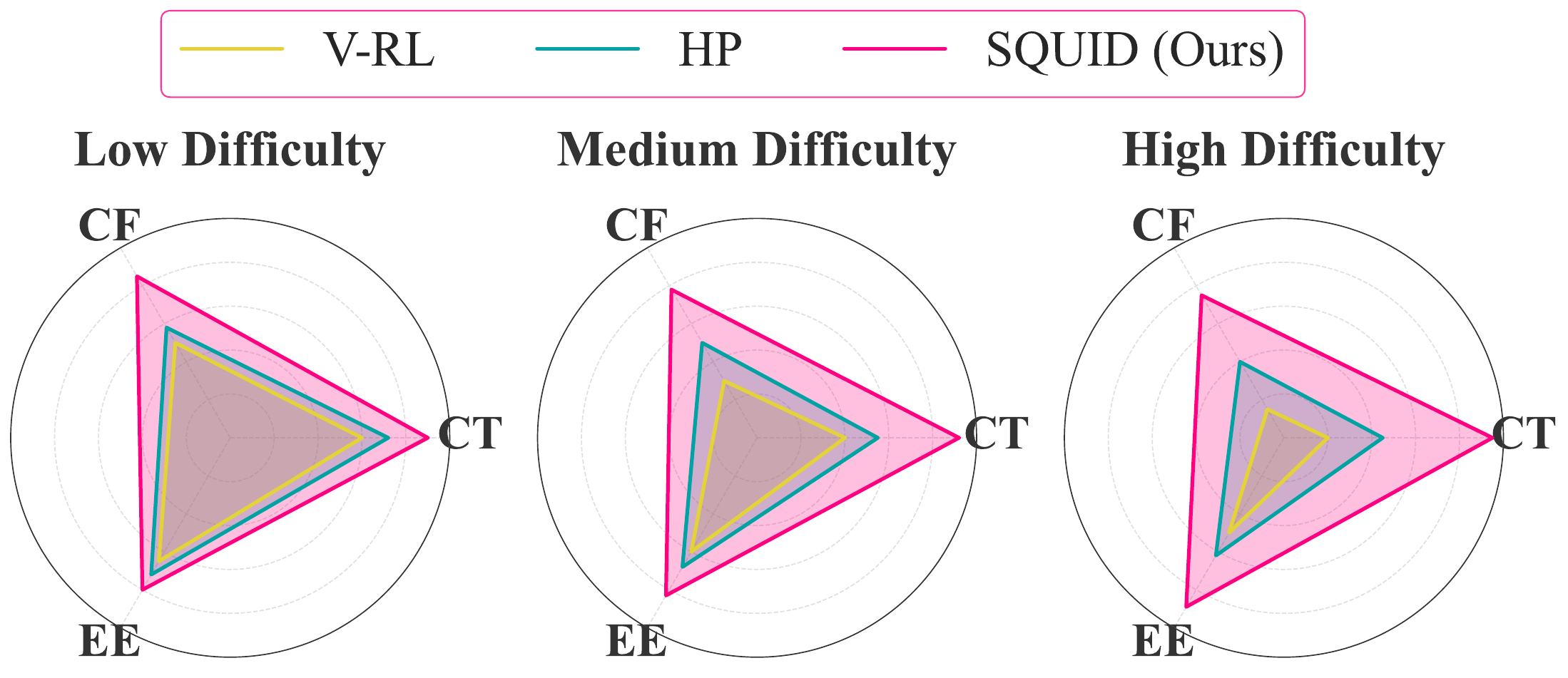}
    \vspace{-5pt}
    \caption{Comparison of Completion Time (CT), Collision Frequency (CF), and Energy Efficiency (EE) across difficulty levels. \textcolor{squidpink}{\textsc{squid}} outperforms \textcolor{squidyellow}{V-RL} and \textcolor{squidgreen}{HP}, maintaining faster traversal, lower collisions, and better energy efficiency, with increasing advantages in higher difficulty tunnels.}
    \label{fig::comparison_baseline}
    \vspace{-8pt}
\end{figure}

\subsubsection{Traversal Efficiency, Collision Avoidance, and Energy Consumption}

The radar plots in Fig.~\ref{fig::comparison_baseline} summarize three key performance metrics—Completion Time, Collision Frequency, and Energy Consumption—across low, medium, and high difficulty levels. These three interdependent aspects directly impact the efficiency and robustness of the locomotion policy. To maintain a consistent interpretation where higher values indicate better performance, these metrics have been inverted in the plot.

Completion Time: \textsc{squid} consistently achieves faster traversal speeds compared to HP, which tends to be overly cautious. While HP ensures stability, it sacrifices speed, leading to long Completion Time. V-RL occasionally stalls in tight spaces, further increasing traversal time. \textsc{squid} strikes an effective balance, maintaining fast but stable motion across varying tunnel geometries.

Collision Frequency: \textsc{squid} exhibits the lowest collision rates, benefiting from its multi-expert knowledge transfer. It effectively maintains clearance from tunnel walls while adapting dynamically to asymmetric structures. HP struggles in tunnels requiring whole-body posture adaptation, while V-RL frequently collides due to erratic foot placements and inadequate control in constrained spaces.

Energy Consumption: \textsc{squid} achieves greater energy efficiency by ensuring smoother gait transitions and reducing unnecessary corrective movements. HP, although stable, expends more energy due to its slow traversal, which increases total energy expenditure. V-RL is the least efficient, consuming excessive energy due to frequent stops, unstable gaits, and inefficient motor commands.

\subsection{Ablation Studies} 

% We conducted a series of ablation experiments to quantify the contribution of each major component in the T–S pipeline:

To systematically assess the contribution of key components in our \textsc{squid} framework, we conduct a series of ablation experiments, selectively modifying critical elements and evaluating their impact on policy performance. The results, summarized in Table IV, report the Success Rates under different ablation settings. These experiments provide insights into the importance of multi-expert learning, procedural generation, privileged perception, and structured reward shaping.

\subsubsection{Single Teacher vs. Multiple Teachers}  
% We trained a student policy from only the best single teacher (the Full Circle tunnel expert) and compared it to the student distilled from all five tunnel-class experts. As shown in Table IV, ST generalizes poorly to non-circular cross-sections, achieving only 70\% success in triangular tunnels at medium difficulty, compared to 88\% with MT. This gap highlights the necessity of multiple specialized teachers.
To test the necessity of multiple expert policies, we train a student policy using only a single teacher, specifically the expert trained in circular tunnels. The resulting policy exhibits moderate success in environments similar to the training distribution but fails to generalize to tunnels with asymmetric constraints, such as triangular or gap tunnels. This highlights that policies trained with a single teacher overfit to specific tunnel geometries, leading to poor adaptability when encountering new structural variations. In contrast, our multi-teacher approach, where each expert specializes in a distinct tunnel class, provides a more diverse knowledge base. The distilled student learns adaptive behaviors across varying tunnel configurations, leading to more consistent success across all settings.

\subsubsection{Procedural Generation Disabled}
% When tunnel shapes are not procedurally varied (i.e., only a single fixed geometry per tunnel class), success rates drop by 5–10\% in the final policy on new unseen configurations. The results confirm that dynamic geometry sampling during training fosters robust adaptation to shape variability.
To examine the role of environmental diversity, we train a student policy in a fixed, non-procedural environment where tunnel shapes remain constant across training episodes. The policy achieves reasonable success in familiar scenarios but fails in tunnels with unexpected variations in curvature, slope, or orientation. Without exposure to procedural variations during training, the policy becomes rigid, adapting poorly to real-world variations. This confirms that procedural generation plays a crucial role in promoting generalization by exposing the policy to a wide range of environmental constraints.

\subsubsection{Two-Layer Elevation Map vs. Ground Elevation Map}

We train teacher policies using a two-layer elevation map that concatenates floor and ceiling elevations and compare them to policies trained with only ground elevation maps. Policies using the two-layer elevation representation fail to converge, as the additional ceiling constraints introduce conflicting optimization objectives, leading to unstable body posture adjustments and frequent stalls. In contrast, policies trained with only ground elevation maps successfully learn stable locomotion strategies, achieving better generalization and traversal efficiency. These results indicate that explicitly encoding ceiling constraints increases learning complexity without improving policy performance, suggesting that alternative ceiling-aware representations should be explored.

Because teacher policies are trained in separate tunnel classes, they can estimate ceiling elevation using only the ground heightmap and adjust posture accordingly. They rely on privileged information to estimate elevation changes, ensuring smooth transitions at tunnel entries and exits. However, distillation with ground elevation alone does not generalize, requiring the student policy to use a depth map for exteroception.

\subsubsection{Reward Shaping Simplifications}
% We tested removing terms from the teacher reward function, such as the vertical velocity penalty (rvp) and the collision penalty (rcoll). Policies trained without collision penalties exhibit significantly higher contact rates with tunnel ceilings, reducing success rates by up to 15\%. Eliminating vertical velocity penalties leads to “hopping” behaviors in half-circle tunnels, further underscoring the importance of reward shaping for stable body posture.
We further analyze how structured reward functions contribute to stable locomotion by removing key components from the expert training stage. Eliminating the collision penalty results in erratic movement patterns, with the policy frequently making contact with tunnel walls and ceilings due to the absence of a strong deterrent against risky postures. Similarly, removing the vertical velocity penalty leads to an increase in destabilizing hopping behaviors, particularly in environments with variable elevation. These behaviors compromise stability and traversal efficiency, underscoring the importance of structured reward shaping for safe and effective locomotion.

%=============================
% TABLE IV: Ablation Results
%=============================
% \begin{table}[ht]
% \centering
% \caption{Ablation study results for medium-difficulty triangular tunnels and high-difficulty bench tunnels. We report success rate (\%) in triangular tunnels and Collision Frequency (collisions/m) in bench tunnels.}
% \label{tab:ablation}
% \begin{tabular}{l c c}
% \toprule
% \textbf{Ablation Setting} & \textbf{Success Rate} & \textbf{Collision Freq} \\
%  & (Triangle, Medium) & (Bench, High) \\
% \midrule
% Full T--S (Ours)          & \textbf{90\%} & \textbf{0.18} \\
% Single Teacher        & 70\%          & 0.36 \\
% No Procedural Gen. (NoPG) & 77\%          & 0.29 \\
% Vision-Only Distilled     & 72\%          & 0.34 \\
% No Collision Penalty      & 65\%          & 0.41 \\
% No Vertical Vel. Penalty  & 68\%          & 0.37 \\
% \bottomrule
% \end{tabular}
% \end{table}   

\begin{table}[ht]
\centering
\begin{tabular}{l c}
\toprule
\textbf{Ablation Setting} & \textbf{Success Rate} \\
\midrule
Full \textsc{squid} (Ours)          & \textbf{70\%} \\
Single Teacher       & 35\%          \\
No Procedural Generation & 39\%          \\
Two-Layer Elevation Map     & 19\%          \\
% No Collision Penalty      & XX\%          \\
% No Vertical Vel. Penalty  & XX\%          \\
\bottomrule
\end{tabular}
\vspace{3pt}
\caption{Success Rates (\%) for different ablation settings, demonstrating the impact of key \textsc{squid} components.}
\vspace{-10pt}
\label{tab:ablation}
\end{table}

% Discussion of Ablations:  
% These ablations underscore that each design choice—from diverse teacher policies to the structured reward function—facilitates robust performance. The largest drop occurs when only a single teacher is used or when collision penalties are removed, indicating that multi-expert knowledge sharing and appropriate reward terms are vital for stable navigation.

\subsection{Real-World Deployment}
To assess the sim-to-real transferability of our \textsc{squid} policy, we deploy the trained model on a Unitree Go2 quadruped in a controlled tunnel testbed (Fig.~\ref{fig::deployment}). The real-world setup consists of modular 1\,m tunnel segments that instantiate multiple constrained geometries at high difficulty, including (i) a narrow circular tunnel with radius $r=0.20$\,m, (ii) an equilateral triangular tunnel with side length $l=0.60$\,m, whose base (floor) is inclined at $40^\circ$, and (iii) a gap tunnel with a $0.50$\,m-wide opening. We varied the real-world tunnel parameters and report the highest difficulty the robot could reliably achieve with limited manual intervention.

The quadruped is equipped with a 4D LiDAR, which generates depth images that are processed through an encoder network before being passed to the deployable \textsc{squid} policy. Unlike simulation, where expert policies can access privileged state information during training, the real-world deployment introduces depth noise, missing returns, providing a challenging test for generalization using only onboard sensing. We evaluate each tunnel configuration for 10 trials. We define a trial as successful if the robot traverses at least 80\% of the tunnel length without requiring external recovery. Under this protocol, the policy achieves a success rate of 70\% in the circular tunnel, 80\% in the gap tunnel, and 60\% in the sloped triangle tunnel, which is the most challenging due to the combined tight clearance and roll perturbation induced by the $30^\circ$ slope. Overall, these results demonstrate that the distilled policy transfers to real hardware and remains functional under realistic sensing artifacts and construction tolerances at high difficulty settings.

% \subsection{Summary of Findings} 

% 1) Our T–S policy outperforms single-policy baselines on all tunnel classes, achieving up to 90\% success rates even at high difficulty.  
% 2) Ablation studies confirm that procedural generation, collision-aware reward shaping, and multi-teacher distillation are crucial for robust behavior.  
% 3) Real-world tests demonstrate that our policy transfers effectively from simulation to physical tunnels, achieving an 82\% success rate under sensor occlusions.

% Collectively, these results verify the effectiveness of our privileged teacher–student pipeline in producing a single locomotion policy that is both adaptive and robust to diverse confined-space geometries.

\section{Conclusions and Future work}
This paper presents \textsc{squid}, a RL framework combining procedural environment generation and privileged policy distillation to achieve robust quadrupedal locomotion in confined 3D tunnel environments. \textsc{squid} leverages multiple expert teacher policies trained on diverse procedurally generated tunnel geometries and distills their specialized knowledge into a unified vision-based student policy, effectively addressing limitations of existing methods such as overspecialization, sensitivity to sensor noise, and reliance on simplified environmental assumptions. Experimental results demonstrate that \textsc{squid} consistently outperforms baseline approaches across various tunnel geometries and difficulty levels, achieving higher success rates, faster traversal times, fewer collisions, and improved energy efficiency. Real-world deployment further validates the robustness of \textsc{squid} under realistic sensor conditions. Our evaluation focuses on structured tunnel-like geometries and does not explicitly model unstructured clutter (e.g., debris, rocks, or irregular protrusions) or tight-turn junction networks. Extending the procedural generator and evaluating in more unstructured confined passages are important directions for future work.

\bibliographystyle{IEEEtran}
\bibliography{references}

\end{document}